\documentclass[letterpaper, 10 pt, conference]{ieeeconf}  % Comment this line out if you need a4paper

\IEEEoverridecommandlockouts                              % This command is only needed if 
\makeatletter
    \let\NAT@parse\undefined
\makeatother
\usepackage[square,numbers,sort&compress]{natbib}
\usepackage{amssymb}
\usepackage{times}
\usepackage{multicol}
\usepackage{amsmath} % assumes amsmath package installed
\usepackage{amssymb}  % assumes amsmath package installed
\usepackage{pifont}
\usepackage{adjustbox}
\usepackage[table, dvipsnames]{xcolor}
\usepackage{graphicx}
\usepackage{xspace}
\usepackage{multirow}
\usepackage{adjustbox}
\usepackage{subcaption}
\usepackage{placeins}
\usepackage{siunitx}
\usepackage[colorlinks = true,
            linkcolor = black,
            urlcolor  = cyan,
            citecolor = magenta,
            bookmarks = true,
            pagebackref]{hyperref}
\let\labelindent\relax % Disables \labelindent already defined error from enumitem
\usepackage{enumitem}
\usepackage{comment}
\usepackage{booktabs}
\usepackage{makecell}

\title{\LARGE \bf
Synthetic vs. Real Training Data for Visual Navigation
}

\author{Lauri Suomela$^{1}$, Sasanka Kuruppu Arachchige$^{1}$, German F. Torres$^{1}$, \\ Harry Edelman$^{2}$, Joni-Kristian Kämäräinen$^{1}$%
\\
{\small $^{1}$Tampere University $^{2}$Turku University of Applied Sciences}%
}

\begin{document}

\newcommand{\best}[0]{\rowcolor{LimeGreen!30}}
\newcommand{\MethodName}[0]{FAINT\xspace}
\newcommand{\MethodNameReal}[0]{FAINT$_{Real}$\xspace}
\newcommand{\MethodNameSim}[0]{FAINT$_{Sim}$\xspace}
\newcommand{\MethodNameSimTen}[0]{FAINT$_{Sim10\%}$\xspace}

\makeatletter
\DeclareRobustCommand\onedot{\futurelet\@let@token\@onedot}
\def\@onedot{\ifx\@let@token.\else.\null\fi}

\def\eg{\emph{e.g}\onedot} \def\Eg{\emph{E.g}\onedot}
\def\ie{\emph{i.e}\onedot} \def\Ie{\emph{I.e}\onedot}
\def\cf{\emph{c.f}\onedot} \def\Cf{\emph{C.f}\onedot}
\def\etc{\emph{etc}\onedot} \def\vs{\emph{vs}\onedot}
\def\wrt{w.r.t\onedot} \def\dof{d.o.f\onedot}
\def\etal{et al\onedot}

\makeatother % Abbreviation definitions

%% Training parameters

% Number of trajectories collected for training
\def\NumTrainTrajectories{\textcolor{red}{X}}
% Number of image pairs in trajectories
\def\NumTrainImagePairs{\textcolor{red}{X}}
% Base learning rate for training
\def\TrainLearningRate{$\num{2e-4}$\xspace} 
% Number of train epochs (Behavior cloning)
\def\NumTrainEpochsBC{\textcolor{red}{X}} 
% Duration of train data collection for BC
\def\BCDataCollectionDuration{12} 
\def\TrainBatchSize{512\xspace}
\def\NumTrainGpus{2}
\def\TrainImageResolution{$224 \times 224$\xspace}
\def\DaggerDecayCoeff{0.8\xspace}

%% Place recognition
\def\PlaceRecognitionImageResolution{$640\times 480$}
\def\PlaceRecognitionFeatureDimension{512}

%% Simulator configuration
% Steps until episode terminated
\def\SimulatorStepLimit{500\xspace}
\def\RouteSuccessThreshold{\SI{0.4}{\meter}\xspace} 
\def\SimulatorStepDuration{\SI{0.25}{\second}\xspace}
% Number of gpus X number of sims per gpu?
\def\NumParallelSims{16\xspace} 
\def\SimRobotRadius{\SI{0.175}{\meter}\xspace}
\def\SimRobotSafetyMargin{\SI{0.2}{\meter}\xspace}
\def\SubgoalMinSpacing{\SI{0.5}{\meter}\xspace}
\def\SubgoalMaxSpacing{\SI{3.0}{\meter}\xspace}

%% Real deployment system configuration
% For both place recognition and goal reaching policy
\def\DeploymentLoopFrequency{\SI{4}{\hertz}}

%% Real robot experiments
\def\NumRealRoutes{10} % Number of navigation test routes for indoor / outdoor general experiment
\def\NumRouteRepetitionsReal{3} % Number route attempts per method
\def\RealReferenceImageInterval{\SI{5}{\second}\xspace} % Such as number of train trajectories, etc.
% The map
\newcommand{\map}{\mathcal{M}}

% Subgoal selection policy
\newcommand{\sgpolicy}{\pi_{s}}

% Goal-reaching policy
\newcommand{\grpolicy}{\pi_{g}}

% Observation image
\newcommand{\obs}{O_t}

% Observation embedding
\newcommand{\zObs}{\mathbf{z}_t}

% Map image at state/node s
\newcommand{\node}{S_{t}}

% Map embedding at state/node s
\newcommand{\zNode}{\mathbf{z}_{s}} % Math symbols

\maketitle
\thispagestyle{empty}
\pagestyle{empty}

%%%%%%%%%%%%%%%%%%%%%%%%%%%%%%%%%%%%%%%%%%%%%%%%%%%%%%%%%%%%%%%%%%%%%%%%%%%%%%%%
\begin{abstract}
This paper investigates how the performance of visual navigation policies trained in simulation compares to policies trained with real-world data.
Performance degradation of simulator-trained policies is often significant when they are evaluated in the real world. However, despite this well-known sim-to-real gap, we demonstrate that simulator-trained policies can match the performance of their real-world-trained counterparts.
Central to our approach is a navigation policy architecture that bridges the sim-to-real appearance gap by leveraging pretrained visual representations and runs real-time on robot hardware.
Evaluations on a wheeled mobile robot show that the proposed policy, when trained in simulation, outperforms its real-world-trained version by 31 and the prior state-of-the-art methods by 50 points in navigation success rate. Policy generalization is verified by deploying the same model onboard a drone.
Our results highlight the importance of diverse image encoder pretraining for sim-to-real generalization, and identify on-policy learning as a key advantage of simulated training over training with real data.
Code, model checkpoints and multimedia materials are available at \href{https://lasuomela.github.io/faint/}{lasuomela.github.io/faint}.
\vspace{-0.2cm}
\end{abstract}

%===============================================================================
\section{Introduction}
\label{sec:intro}

Recently, learning-based visual navigation methods have received attention as potential replacements for traditional sense-plan-act approaches that leverage geometric environment representations.
Navigation systems with learned components have many advantages, for example, being able to utilize semantic information to infer traversability~\citep{kahn_badgr_2021, castro_how_2023} and guide exploration~\citep{mousavian_visual_2019, gervet_navigating_2023}, and incorporate other task specifications in addition to metric coordinates~\citep{anderson_evaluation_2018}.
The primary robot learning paradigms involve imitation learning (IL) from real-world robot datasets~\cite{kahn_badgr_2021, shah_ving_2021, shahGNMGeneralNavigation2023, shahViNTLargeScaleMultiTask2023, sridhar_nomad_2024, yang_pushing_2024}, and reinforcement learning (RL) or imitation from scripted~\cite{loquercio_learning_2021, ehsani_spoc_2024} or human experts~\cite{ramrakhya_habitat-web_2022, gervet_navigating_2023} in simulation~\cite{hoeller_learning_2021, truong_rethinking_2023, kulkarni_reinforcement_2024, geles_demonstrating_2024}. The performance of learned robot policies heavily depends on the quality, quantity, and diversity of training data~\citep{hu_data_2024}.
Real and synthetic data collection each have distinct strengths and weaknesses with respect to these factors.
Real-world data collection is labor-intensive and platform-specific, yet it reduces domain shifts at policy deployment, while synthetic data generation offers greater scalability and flexibility but faces the \emph{simulation-to-real} (sim2real) gap~\citep{hofer_sim2real_2021}, limiting generalization to real environments.

\begin{figure}[!t]
    \begin{center}
        \includegraphics[trim={0cm 0 0 0cm}, clip, width=0.99\linewidth]{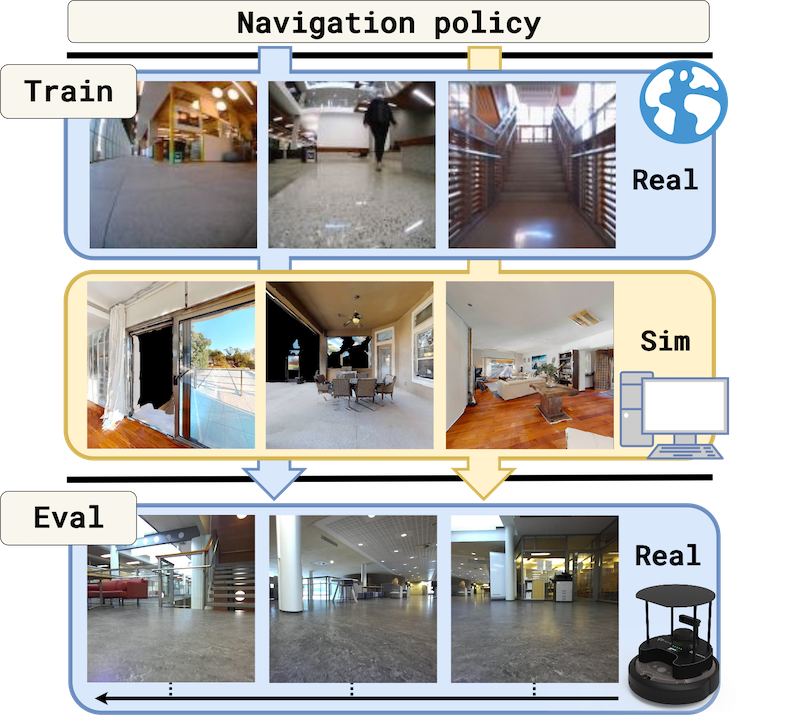}
        \captionof{figure}{We investigate how simulation-trained navigation policies compare to ones trained with real-world data when deployed on a real robot.
        }
        \label{fig:opener}
    \end{center}
    \vspace{-0.5cm}
\end{figure}

Our work investigates how navigation performance differs between policies trained with real-world data and those trained entirely in simulation. 
Although prior works~\cite{kadian_sim2real_2020, sadek_-depth_2022, gervet_navigating_2023, silwal_what_2024} have assessed the sim2real gap by evaluating simulation-trained policies in simulated and real environments, they have not directly compared these policies with those learned from real-world data.
We address this gap by evaluating whether simulation-only training can achieve performance competitive with real-world data.

Previously proposed navigation policies have been tailored for either synthetic or real-world data. To facilitate our experiments, we introduce a novel policy architecture for visual topological navigation: 
\textbf{F}ast
\textbf{A}ppearance-\textbf{I}nvariant
\textbf{N}avigation
\textbf{T}ransformer
(\MethodName). \MethodName can be trained on either real or simulated data, is lightweight enough for deployment on resource-constrained robot hardware, and demonstrates robust sim-to-real transfer.
We deployed sim and real-data-trained versions of the policy on a wheeled mobile robot and a drone, and performed more than 50 hours of evaluation in challenging real-world indoor environments.

Our findings demonstrate that a navigation policy trained entirely in simulation can perform on par with—or even outperform—those trained on real-world data.
Simulation-trained \MethodName surpasses its real-world-trained counterpart by 31 points and the previous state-of-the-art by 50 points in navigation success rate.
Despite being trained with fully synthetic data, the policy can successfully adapt to unseen environment conditions and different robot platforms.
Based on these results, we identify scalable data generation and the ability to perform on-policy learning as key advantages of simulation over real-world data.

%===============================================================================
\section{Related work}
\label{sec:related}
\noindent\textbf{Learning from real-world data.}
Most works formulate learning navigation from real-world demonstrations as a goal-conditioned imitation learning~\citep{codevilla_end--end_2018} problem.
The main challenges are related to generalization across unseen environments and embodiments.
\citet{kahn_badgr_2021} train a navigation model in one of the first works to demonstrate generalization to novel environments.
Shah~\etal~\citep{shahGNMGeneralNavigation2023, shahViNTLargeScaleMultiTask2023, sridhar_nomad_2024} achieve generalization across robot platforms by training topological navigation models with data collected from multiple different robots.
\citet{suomela_placenav_2024} improve upon this line of work by reducing the dependency on robot-originated training data through the use of place recognition models~\citep{masone_survey_2021}.
We build on these prior architectures, but introduce improvements that we find critical for sim2real generalization.

\vspace{0.2cm}
\noindent\textbf{Learning from synthetic data.}
Simulation offers scalability and tailorability, making it attractive for training navigation policies. However, the sim2real gap~\citep{hofer_sim2real_2021} remains a major challenge for real-world deployment~\citep{gervet_navigating_2023}. While policies trained on raw RGB inputs with domain randomization have shown some success, their generalization remains limited~\citep{zhu_target-driven_2017, sadeghi_cad2rl_2017, meng_scaling_2020}.
Training policies on sensor abstractions such as depth images~\citep{loquercio_learning_2021, truong_indoorsim--outdoorreal_2024}, segmentation masks~\citep{mousavian_visual_2019, geles_demonstrating_2024}, and feature points~\citep{kaufmann_deep_2020} is a promising alternative.
Recently, feature maps from \emph{pre-trained visual representation} (PVR) models have proven effective for bridging the sim2real gap. Ehsani~\etal~\citep{ehsani_spoc_2024, zeng_poliformer_2024, hu_flare_2024, eftekhar_one_2024} use a frozen SigLIP~\citep{zhai_sigmoid_2023} encoder for navigation and manipulation tasks, while \citet{silwal_what_2024} demonstrate successful sim2real transfer in image-goal navigation using a VC-1~\citep{majumdar_where_2024} encoder.
However, these large PVR's are unsuitable for real-time deployment on resource-constrained robots.
In this work, we show that even smaller, distilled PVR models~\citep{shang_theia_2024} enable sim2real transfer while being suitable for on-robot execution.

\vspace{0.2cm}
\noindent\textbf{Sim2Real investigations.}
Substantial amounts of work have been put into studying and quantifying the sim2real gap by comparing policies' performances in simulation and the real world~\citep{kadian_sim2real_2020, sadek_multi-object_2023, truong_rethinking_2023, gervet_navigating_2023}.
What we identify as still missing is a real-world comparison of the performance of navigation models trained with synthetic and real datasets.
The work most similar to ours is the investigation by \citet{silwal_what_2024}, which examines the performance of manipulation policies trained with real and simulated data in real-world settings. They find that simulated policies trained with few-shot imitation learning exhibit poor sim2real transfer and underperform compared to real-data policies. However, they also train an image-goal navigation policy using large-scale reinforcement learning in simulation, which performs well in real-world environments. Notably, they do not compare the image-goal policy to a model trained on real-world data, likely because there are no suitable real-world datasets. In contrast, our investigation focuses on visual topological navigation, a task for which real-world datasets are available. This allows us to directly compare the policies trained on real and synthetic data for the same task.

%===============================================================================
\section{Methods}
\label{sec:methods}

% Model architecture
\begin{figure*}[h]
    \centering
    \begin{adjustbox}{width=0.80\linewidth}
    \includegraphics[trim={0.2cm 0 0cm 0},clip, width=0.9\linewidth]{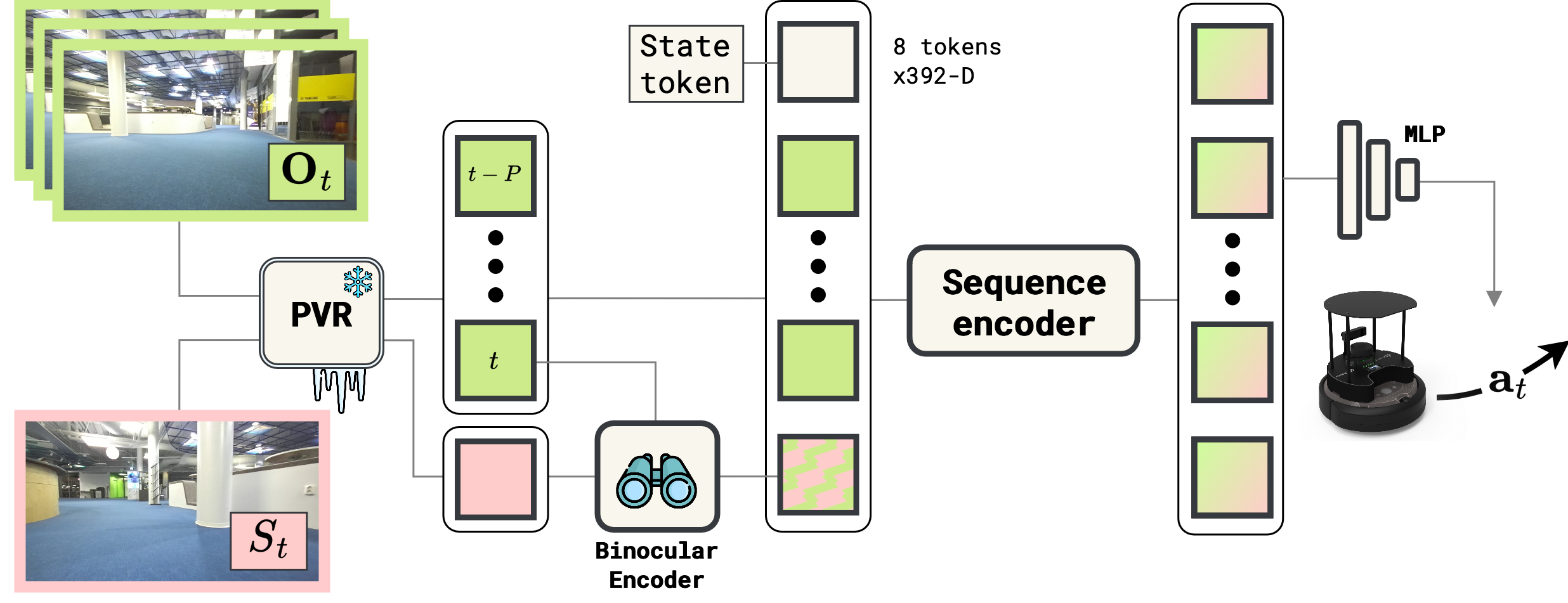}
    \end{adjustbox}
    
    \caption{\textbf{Model architecture.} \MethodName implements the goal-reaching policy $\mathbf{a}_t = \grpolicy (\mathbf{O}_t, \node)$.
    Observation and subgoal images are encoded with a frozen PVR, and a binocular encoder refines the goal tokens by conditioning on the latest observation. A sequence encoder with a predictor head then produces the actions $\mathbf{a}_t$.
    Subgoals $S_t$ are obtained from a separate subgoal selection policy $\sgpolicy$.
    }
    \label{fig:model_architecture}
    \vspace{-0.5cm}
\end{figure*}

In this Section, we present the definition of the navigation task (Sec.~\ref{sec:problem_formulation}), outline the architecture of the model used in the experiments (Sec.~\ref{sec:model_arhitecture}), and describe the datasets and learning approach for synthetic (Sec.~\ref{sec:learning_in_sim}) and real-world (Sec.~\ref{sec:learning_in_real}) data.

\subsection{Problem formulation}
\label{sec:problem_formulation}

\noindent\textbf{Topological visual navigation.}
We perform the investigation in the context of topological image-goal navigation~\citep{savinov_semi-parametric_2018, shahGNMGeneralNavigation2023, suomela_placenav_2024} because it is a navigation task with some of the most extensive real-world datasets available.
Topological approaches divide a navigation route into a set of intermediate subgoals 
$\left\{ s_0,s_1,\ldots,s_n \right\}$, each with an associated image observation. These subgoals comprise a topological map $\map$, created from images collected prior to robot deployment. During navigation, at each time step $t$ a \emph{subgoal selection policy} $\sgpolicy$ finds the next subgoal $s_t$ along the route to the final goal, and returns the corresponding subgoal image $\node$:
\begin{equation}
    \node = \sgpolicy (\obs,\map)
\end{equation}
where $\obs$ is the current observation image.
Given $\node$ and a sequence of $P$ recent observations $\mathbf{O}_t = \left\{ O_{t-P+1}, \ldots, O_{t} \right\}$, a \emph{goal-reaching policy} $\grpolicy$ then produces a sequence of $H$ robot control commands $\mathbf{a}_t = \left\{ a_t, \ldots, a_{t+H-1} \right\}$ towards the subgoal:
\begin{equation}
    \mathbf{a}_t = \grpolicy (\mathbf{O}_t, \node).
\end{equation}

We adopt the subgoal selection method from~\citet{suomela_placenav_2024}, and perform the selection with place recognition~\citep{masone_survey_2021} models that can be trained with large-scale datasets from~\eg~Google Streetview~\citep{berton_rethinking_2022}. This lets us focus on the goal-reaching policies which are more tightly tied to the robot embodiment and for which the training data is scarce.

\vspace{0.2cm}
\noindent\textbf{Action space.}
The goal-reaching policy outputs trajectory \emph{waypoints} relative to the robot coordinate frame, as it allows embodiment-agnostic control and direct comparison to relevant prior methods~\citep{shahGNMGeneralNavigation2023, shahViNTLargeScaleMultiTask2023, suomela_placenav_2024}.
Each waypoint $a \in \mathbf{a}_t$ is a pose $a = [x, y, \theta] \in SE(2)$ where $(x, y)$ is the position and $\theta$ the orientation. During deployment, a simple PD-controller estimates velocity commands from the waypoints.

\subsection{\MethodName model architecture}
\label{sec:model_arhitecture}

We improve prior goal-reaching policies~\citep{shahViNTLargeScaleMultiTask2023} by integrating a \emph{pretrained visual representation} (PVR's) and a novel \emph{binocular goal encoder}.
\MethodName has just 12M parameters—half the size of prior models~\citep{sridhar_nomad_2024}—enabling real-time inference on resource-constrained hardware.
An overview of the architecture is shown in Figure~\ref{fig:model_architecture}. We describe each component in detail below.

\noindent\textbf{Pretrained visual representation.}
As a key to bridging the sim-to-real appearance gap, we leverage image encoders pretrained on a diverse visual tasks.
We adopt the $5M$-parameter \emph{Tiny CDDSV} variant of the Theia encoder~\citep{shang_theia_2024}, which distills representations from CLIP~\citep{radford_learning_2021}, DiNOv2~\citep{oquab2024dinov}, Depth Anything~\citep{yang_depth_2024}, Segment Anything~\citep{kirillov_segment_2023}, and ViT~\citep{dosovitskiy_image_2021}.
Despite its small size, it demonstrates robust sim-to-real generalization.
The weights are frozen during training to avoid overfitting.

\begin{figure}[!b]
    \centering
    \begin{adjustbox}{width=0.9\linewidth}
    \includegraphics[trim={0 0 0 0},clip, width=1.0\linewidth]{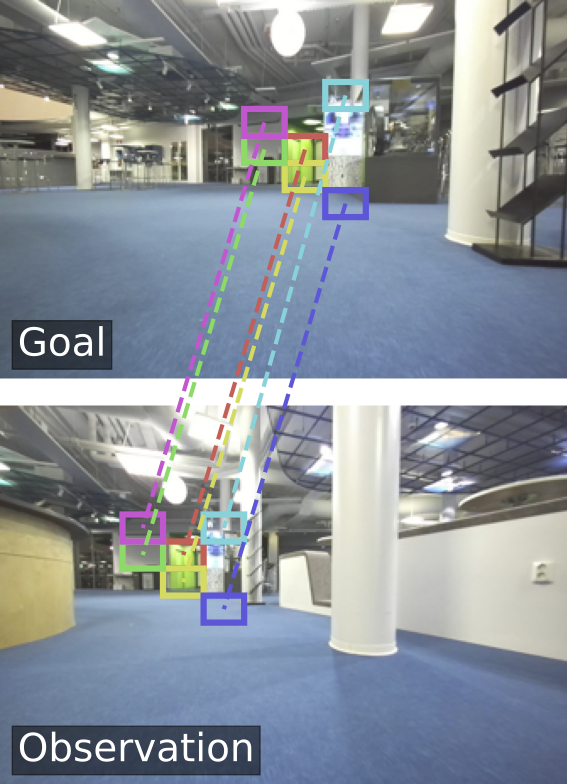}
    \includegraphics[trim={0 0 0 0},clip, width=1.0\linewidth]{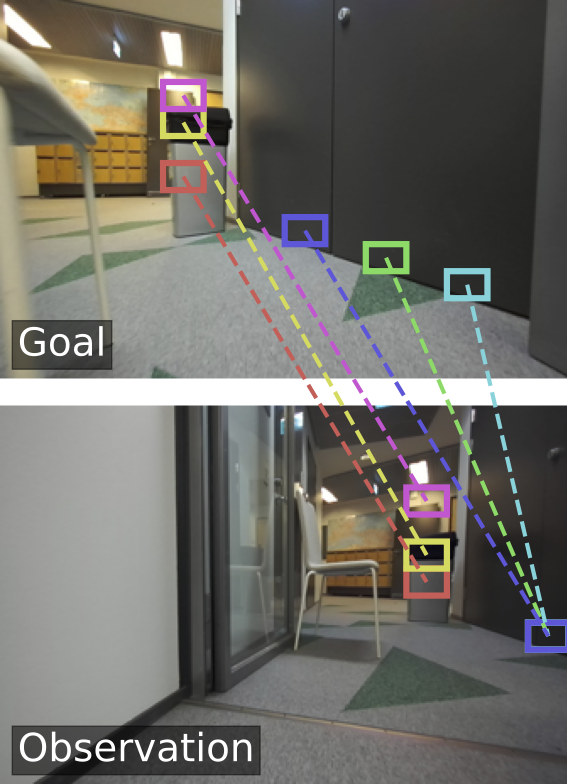}
    \end{adjustbox}
    \caption{Implicit correspondences of the six highest attention values in the binocular encoder's first cross-attention layer.}
    \label{fig:obs_goal_attn}
\end{figure}

\noindent\textbf{Binocular encoder.}
Previous work~\citep{shahViNTLargeScaleMultiTask2023, sun_fgprompt_2023} has shown that conditioning the goal image on the current observation improves navigation performance, for instance by facilitating estimation of the relative pose between the robot and the goal~\citep{bono_end--end_2023}.
However, standard methods that concatenate the observation and goal images along the channel dimension are incompatible with pretrained image encoders.
Inspired by the binocular vision architecture of~\citet{weinzaepfel_croco_2022}, we utilize a transformer decoder to extract correspondences between the encoded observation and goal tokens.
More specifically, the decoder alternates between self-attention on the goal tokens and cross-attention on the observation tokens.
A key strength of this approach is its ability to learn navigation-relevant cues directly from arbitrary, frozen pretrained embeddings. As illustrated in Fig.~\ref{fig:obs_goal_attn}, our binocular encoder identifies matches between image features despite being trained end-to-end with the rest of the policy, without explicit supervision. We use 4 transformer layers with 4 attention heads.
%

% Binocular decoder attention viz

\vspace{0.15cm}
\noindent\textbf{Sequence encoder.}
The observation and conditioned goal tokens are processed by a transformer encoder with non-causal self-attention.
Before input into the sequence encoder, the patch tokens of each image are compressed into a one-dimensional vector with a 2D convolution layer followed by flattening, similar to~\cite{majumdar_where_2024}. A learnable state token is added to the sequence, and its corresponding token from the sequence encoder output is passed to a predictor head. This produces the final output, a sequence of waypoints $\mathbf{a}_{t}$. The sequence encoder consists of 4 layers with 4 attention heads each.

\subsection{Learning to navigate in simulation}
\label{sec:learning_in_sim}

\begin{figure}[t]
    \centering
    \includegraphics[trim={0cm 0cm 0.55cm 0.5cm }, clip, width=0.9\linewidth]{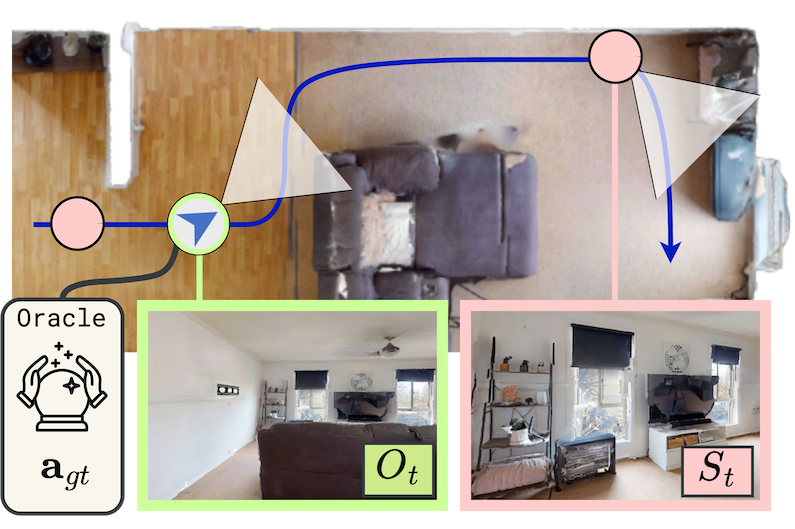}
    \caption{Training data collected from the simulator - oracle actions $\mathbf{a}_{gt}$ that control the agent, agent observation $O_{t}$, and subgoal image $ S_t$.}
    \label{fig:learning_approach}
\end{figure}

\noindent\textbf{Learning goal-reaching from a shortest-path oracle.}
We train a learning-based agent to mimic a scripted oracle agent that follows the shortest paths between the episode start and goal locations.
The oracle has privileged access to the simulator state and utilizes a proportional controller for path tracking.
The student agent is trained as if the oracle was navigating a sequence of subgoals $s \in \map$ along the shortest path.
The agent predicts the oracle actions $\mathbf{a}_{gt}$ while only having access to the $P=6$ latest egocentric camera observations $\mathbf{O}_{t}$, and the subgoal image $S_t$ captured at the next subgoal pose $s_t$.
The predicted actions are trained to minimize the \emph{mean squared error} loss
$
\mathcal{L} = MSE(\mathbf{a}_{t}, \mathbf{a}_{gt})
$.
The set of subgoals $\map$ is randomly sampled along the shortest path so that for each consecutive subgoal the geodesic distance $d(s_n, s_{n+1}) \in [d_{min}, d_{max}]$, where $d_{min}=\SI{0.5}{\meter}, d_{max}=\SI{3.0}{\meter}$ are the minimum and maximum subgoal separation. The subgoal at the time step $t$ is chosen from $\map$ based on the distance from the position of the agent. At each time step, the oracle agent is rolled out for $H=5$ steps to acquire future actions $\mathbf{a}_{gt}$ for each pair of observation and subgoal.
Each triplet $(\mathbf{O}_{t}, S_t, \mathbf{a}_{gt})$, illustrated in Fig.~\ref{fig:learning_approach}, is saved to disk for use as training data for the student agent.

% Sim data generation illustration    

\noindent\textbf{Data distribution.}
Naively imitating the oracle agent leads to poor performance, as shown in Sec.~\ref{sec:sim2real_ablation}. During deployment, compounding prediction errors lead to covariate shift between training data and actual observations~\cite{spencer_feedback_2021}. If a policy has only been trained with shortest-path trajectories, it might not be able to recover back after ending in a state outside the shortest-path distribution. Thus, we employ \emph{DAgger}~\cite{ross_reduction_2011} to diversify the state distribution of the training data. During training data collection, the simulated agent executes the student policy action instead of $\mathbf{a}_{gt}$ with probability
    $p(\mathbf{a}_{t}) = \beta^{r}$,
where $\beta$ is a decay coefficient and $r$ is the number of the current training round. The student action is only executed if it does not lead to a collision.

\noindent\textbf{Simulator \& training setup.}
Training was carried out in the Habitat~\cite{savva_habitat_2019} simulator with the train split of the HM3D environments and the PointNav route dataset~\cite{ramakrishnan_habitat-matterport_2021}.
We sampled routes where the agent can reach the goal within \SimulatorStepLimit steps without collisions. Agent radius was set to \SI{0.1}{\meter} and its movement was simulated by kinematic control~\citep{truong_rethinking_2023}. The RGB camera \emph{field-of-view} (FOV) was set to \SI{110}{\degree} with resolution of $224 \times 126$.
We trained models for 10 rounds of DAgger with $\beta=0.8$,  batch size $512$, AdamW~\cite{loshchilov_decoupled_2019} optimizer and initial learning rate \TrainLearningRate, decayed with cosine schedule. Images were augmented with color jitter and posterization.

\subsection{Learning from real-world data}
\label{sec:learning_in_real}

The real-world-data version of \MethodName was trained with the publicly available topological navigation datasets, specifically RECON~\citep{shahRapidExplorationOpenWorld2022}, GoStanford~\citep{hiroseDeepVisualMPCPolicy2019}, SACSoN~\citep{hirose_sacson_2024}, SCAND~\citep{karnan_socially_2022}, and TartanDrive~\citep{triestTartanDriveLargeScaleDataset2022}. The synthetic part of GoStanford was omitted to avoid mixing the real and synthetic data. Trajectories, when sampled at \SI{4}{\hertz}, have $\sim1.2M$ image frames.
We follow the training procedure described by~\citet{shahViNTLargeScaleMultiTask2023} with the difference that we omit the temporal distance prediction.
To produce training data, pairs of observation and goal images are sampled from the dataset trajectories. After sampling a sequence of observations $\mathbf{O}_{t}$, a goal image $S_t$ is picked randomly from the same trajectory, $[l_{min}, \ldots, l_{max}]$ frames in the future from $t$, similar to hindsight relabeling~\citep{ghosh_learning_2021}.
The $H$ future poses $\mathbf{a}_{gt}$ relative to the current pose of the robot are then used as action labels for training. To enable learning across data from heterogeneous robots, the waypoints are normalized by the average waypoint distance of each dataset~\citep{shahGNMGeneralNavigation2023}.
We used the same training setup as with simulated training, except a smaller batch size of 256 and initial learning rate $5 \times 10^{-4}$.

%===============================================================================
\section{Experiments}
\label{sec:experiments}
% Test route types
\begin{figure*}[t]
    \centering
    \begin{adjustbox}{width=1.0\linewidth}
    \begin{subfigure}[t]{0.45\textwidth}
        \centering
        \includegraphics[trim={5cm 0 5cm 0}, clip, width=\textwidth]{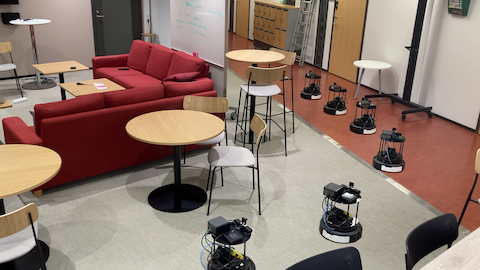}
        \caption{Open space}
        \label{fig:image1}
    \end{subfigure}
    \hfill
    \begin{subfigure}[t]{0.45\textwidth}
        \centering
        \includegraphics[trim={5cm 0 5cm 0}, clip, width=\textwidth]{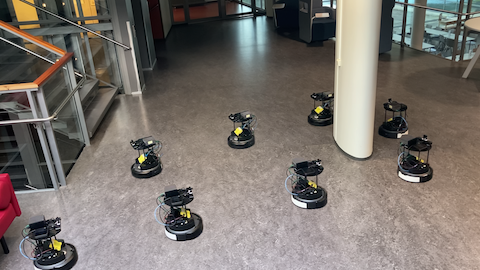}
        \caption{Tight maneuver}
        \label{fig:image2}
    \end{subfigure}
    \begin{subfigure}[t]{0.45\textwidth}
        \centering
        \includegraphics[trim={5cm 0 5cm 0}, clip, width=\textwidth]{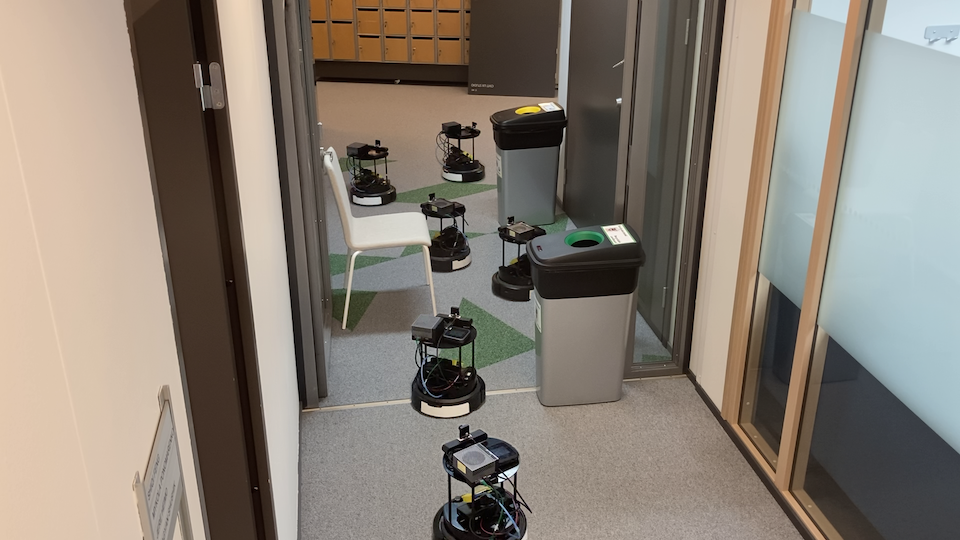}
        \caption{Clutter}
        \label{fig:image3}
    \end{subfigure}
    \end{adjustbox}
    \caption{Example segments from different types of test routes.}
    \label{fig:test_env_types}
    \vspace{-0.3cm}
\end{figure*}

We conducted real-world navigation tests in various indoor environments in experiments designed to answer the following questions.

\begin{itemize}[topsep=0pt, partopsep=0pt, parsep=1pt, itemsep=0pt, leftmargin=0pt]

    \item \textbf{Q1}: How well can a policy perform when trained in simulation instead of with real-world data?

    \item \textbf{Q2}: How does \MethodName trained with synthetic data compare to the previous state-of-the-art?
    
    \item \textbf{Q3}: How do different architectural choices affect the policy's sim-to-real generalization?

    \item \textbf{Q4}: Does training in simulation require the deployment embodiment to closely match the training embodiment?

\end{itemize}

\subsection{General setup}
\label{sec:exp_gen_setup}
\noindent\textbf{Hardware.}
Experiments \textbf{Q1.}-\textbf{Q3.} were performed on a Turtlebot4 robot with a \SI{110}{\degree} FOV ZED 2i camera and an Nvidia Jetson Orin AGX, moving at \SI{0.3}{\metre/\second}.
For studying \textbf{Q4.}, \MethodName was deployed on a custom-built \emph{Agipix}-drone, equipped with a forward-facing \SI{110}{\degree} FOV USB camera and a Jetson Orin NX. All computation was performed on board the robots.

\noindent\textbf{Deployment.}
\MethodName was deployed within the PlaceNav~\citep{suomela_placenav_2024} framework, which divides navigation into separate goal-reaching and subgoal selection policies.
Subgoal selection was performed with a ResNet18~\citep{he_deep_2016} variant of the EigenPlaces~\citep{berton_eigenplaces_2023} place recognition model followed by Bayesian filtering.
The subgoal selection and goal-reaching policies run in separate threads, both at \DeploymentLoopFrequency.
To navigate a route, the robot is first teleoperated to capture map images. A new image is added to the topological map every \SI{5}{\second}, meaning node spacing of $\sim$\SI{1.5}{\metre} at robot speed of \SI{0.3}{\metre/\second}.
During navigation, the robot follows the sequence of image goals from the beginning to the end of the route.

\noindent\textbf{Evaluation.}
We evaluated navigation performance by average \emph{success rate} (SR)~\citep{anderson_evaluation_2018}.
Similar to~\citep{suomela_placenav_2024}, a real-world navigation episode is considered successful if the navigation system’s subgoal selection module localizes to the last image of the topological map.
An episode is unsuccessful if the robot collides with the environment or gets lost in such a way that it cannot return to the test route. In the simulator, we consider an episode successful if the agent arrives within \SI{0.4}{\meter} of the goal within 500 simulator steps.

\noindent\textbf{Test environments.}
The experiments were carried out in various indoor environments including a real apartment, offices, public spaces on a university campus, and a nuclear fallout shelter. The tests were limited to indoor environments to reduce the domain gap between the deployment and the simulated training environments.
The test routes had different features relevant to navigation, illustrated in Figures~\ref{fig:test_env_types} and~\ref{fig:control_exp_illuminatinons_example}.
The lengths of the test routes ranged from 5 to 25 meters.

\subsection{Synthetic vs. real training data}
\label{sec:sim_vs_real}

% Train steps: Sim ~30k, Real ~90k (Half batch size), Sim 10x ~110k
To answer \textbf{Q1.}, we trained versions of \MethodName with different amounts of real and synthetic data and compared the models' performances in the real-world.
\MethodNameReal was trained with behavior cloning (BC), the simulated ones with DAgger.  

\noindent\textbf{Results.}
Table~\ref{tab:real_vs_sim_comp} shows the experiment results divided by route category.
When training with the same number of samples, real-world data produces better navigation performance than the synthetic data.
On the 'open space'-routes, the performance is similar, but on the more challenging routes
\begin{table}
    \centering
    \caption{\textbf{\MethodName} trained with \textbf{real vs. simulated} data, \emph{success rates} ($\uparrow$) over 3 repetitions of 7 routes per category.}
    \label{tab:real_vs_sim_comp}
    \begin{adjustbox}{width=\linewidth}
    
\begin{tabular}{lcccc|c}
         %\multirow{2}{*}{\textbf{Method}} &
         \multirow{2}{*}{\textbf{Dataset}} & \multirow{2}{*}{\textbf{Samples}} & \makecell{\textbf{Open} \\ \textbf{space}} & \makecell{\textbf{Tight} \\ \textbf{maneuver}} & \textbf{Clutter} & \textbf{Total} \\ 

 &   & $n=21$ & $21$ & $21$ & $63$ \\ \hline
         
 %\MethodName &
 Real & $1.2M$ & $0.43$ & $0.52$ & $0.38$ & $0.44$ \\

 %& 
 Sim$_{10\%}$ & $1.2M$ & $0.57$ & $0.05$ & $0.05$ & $0.22$ \\
  
 %\best &
 \best Sim & $12.0M$ & $\mathbf{0.86}$ & $\mathbf{0.62}$ & $\mathbf{0.76}$ & $\mathbf{0.75}$ \\
 \hline
\end{tabular}
        
    \end{adjustbox}
\end{table}
the $Sim10\%$ policy often collides as result of cutting too close to obstacles.
A 10-fold increase in the amount of synthetically generated data, however, leads to a drastic performance increase. \MethodNameSim outperforms the other two by a wide margin in all categories.
It is able to perform complex maneuvers to~\eg~go around obstacles, and reach goal images not within the immediate camera view.

Some of the performance gap between $Real$ and $Sim$ may be explained by dataset size—\MethodNameReal trained on 12M samples might match \MethodNameSim.
However, scaling simulated data is essentially free, while real-world data collection is very labor-intensive.
We suggest on-policy learning as a more insightful explanation. \MethodNameSim trained with BC (see Sec.~\ref{sec:sim2real_ablation}) performs poorly despite being trained with $12M$ samples, demonstrating that scale alone is not sufficient.
Results in~\citep{de_haan_causal_2019} show that DAgger can require a high number of expert queries to work properly. We hypothesize this to be a partial cause of the performance gap between $Sim10\%$ and $Sim$, not the difference in data scale per se.
Interestingly, \MethodNameReal exhibits similar failure modes as \MethodNameSim trained with BC, getting stuck in feedback loops such as spinning in place~\citep{spencer_feedback_2021}.
This parallel suggests that the behavior cloning's inability to handle compounding errors also affects real-world policies trained without on-policy corrections.
This underscores the potential of simulation for robot learning, as large-scale on-policy learning is impractical in the real-world.

\subsection{SOTA comparison}
\label{sec:sota_comp}
\begin{table}
    \centering
    \caption{\textbf{ SOTA comparison} \emph{success rates} ($\uparrow$) over 3 repetitions of 4 routes per category.}
    \label{tab:sota_experiment}
    \begin{adjustbox}{width=\linewidth}
    
\begin{tabular}{lcccc|c}
         \multirow{ 2}{*}{\textbf{Method}} & \makecell{\textbf{Open} \\ \textbf{space}} & \makecell{\textbf{Tight} \\ \textbf{maneuver}} & \textbf{Clutter} & \makecell{\textbf{Illumination} \\ \textbf{change}} & \textbf{Total} \\ 

 &  $n=12$ & $12$ & $12$ & $12$ & $48$ \\ \hline
         
 NoMAD~\citep{sridhar_nomad_2024} & $0.08$ & $0.08$ & $0.25$ & $0.00$ & $0.10$ \\

 PlaceNav~\citep{suomela_placenav_2024} & $0.67$ & $0.25$ & $0.33$ & $0.00$ & $0.31$ \\
 
 ViNT~\citep{shahViNTLargeScaleMultiTask2023} & $\mathbf{0.92}$ & $0.25$ & $0.42$ & $0.00$ & $0.40$ \\
  
 \best \MethodName (ours) & $\mathbf{0.92}$ & $\mathbf{0.92}$ & $\mathbf{0.75}$ & $\mathbf{1.00}$ & $\mathbf{0.90}$ \\
 \hline
\end{tabular}
        
    \end{adjustbox}
\end{table}
\begin{figure}[!b]
    \centering
    \begin{adjustbox}{width=1.0\linewidth}
    \includegraphics[trim={0.4cm 0 -1.0cm 0}, clip, width=\linewidth]{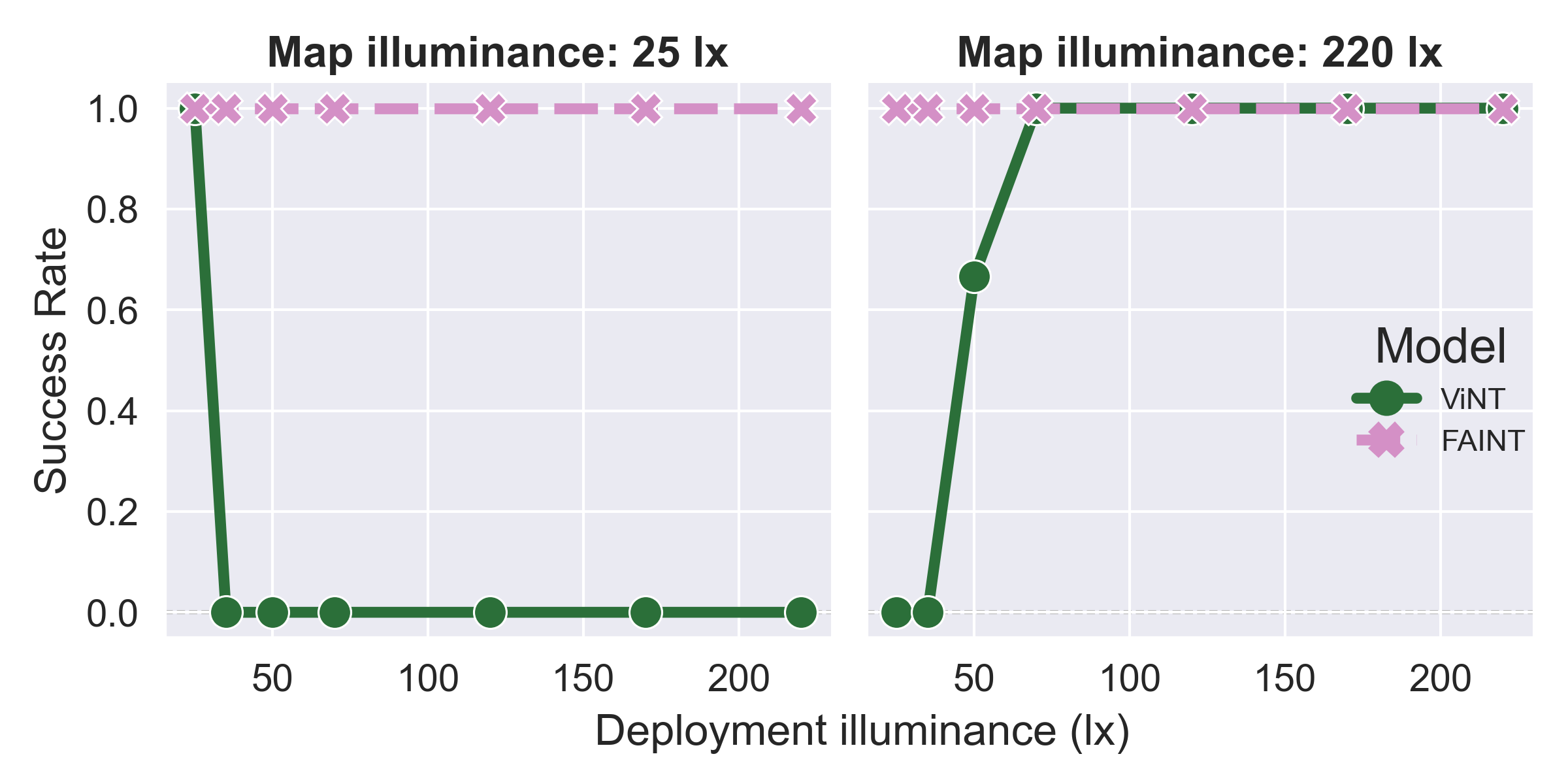}
    \end{adjustbox}
    \begin{adjustbox}{width=1.0\linewidth} %
    \includegraphics[width=\linewidth]{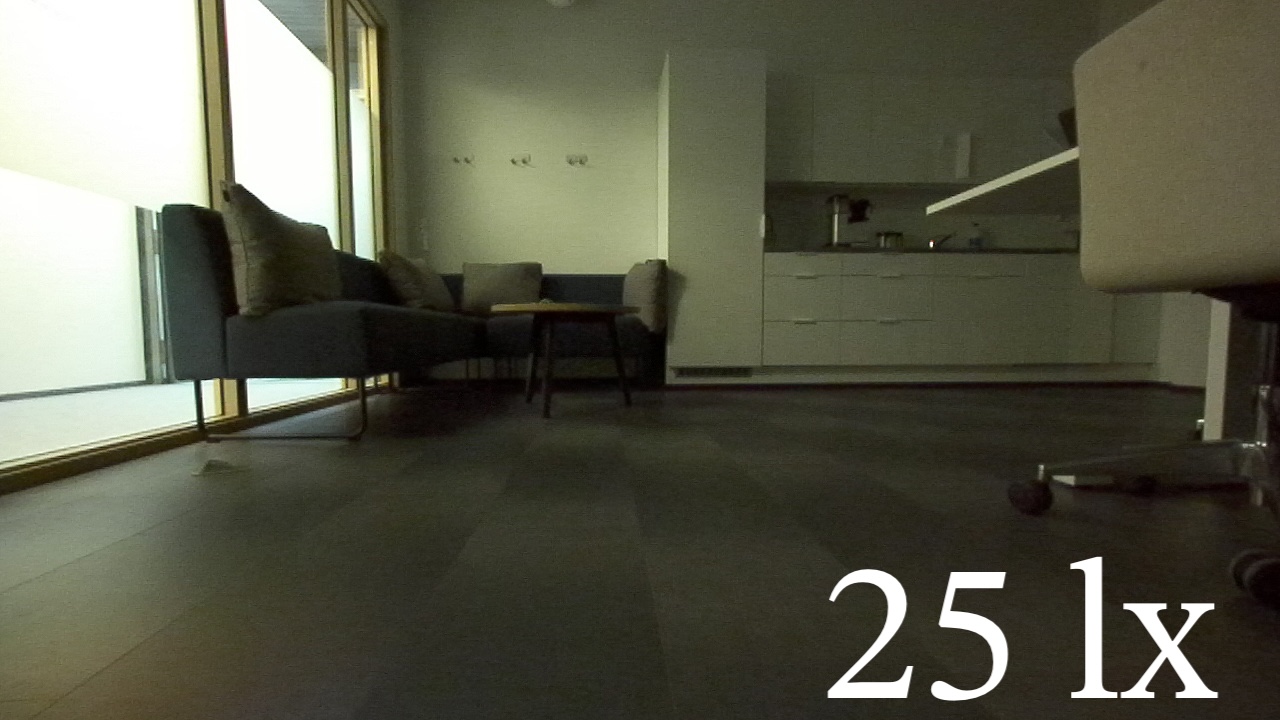}
    \includegraphics[width=\linewidth]{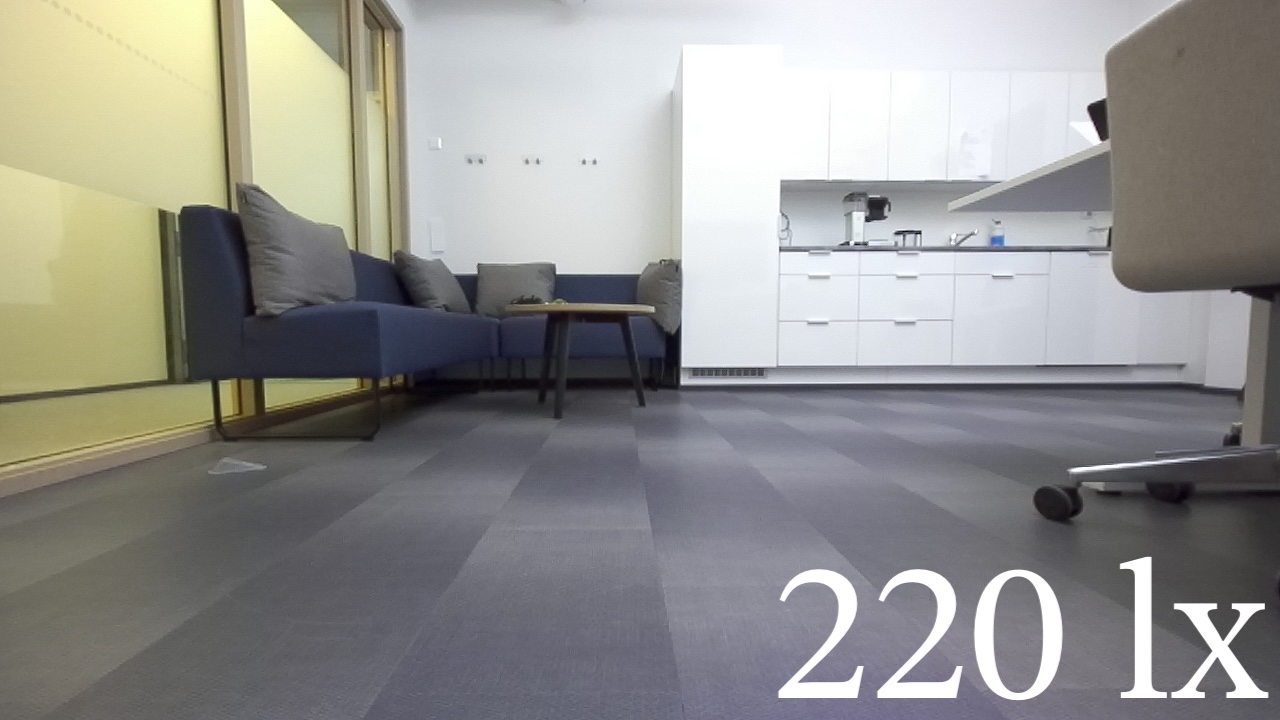}
    \end{adjustbox}
    \caption{ViNT and \MethodName were tested under various illumination levels, with two maps captured under \SI{25}{\lux} and \SI{220}{\lux}.
    }
    \label{fig:control_exp_illuminatinons_example}
\end{figure}

\begin{figure*}[!h]
    \centering
    \includegraphics[
    trim={3cm 5cm 3cm 3cm}, clip,
    width=0.45\linewidth]{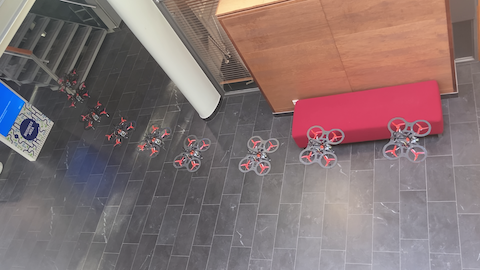}
    \includegraphics[
    trim={0cm 0.3cm 0cm -0.4cm}, clip,
    width=0.49\linewidth]{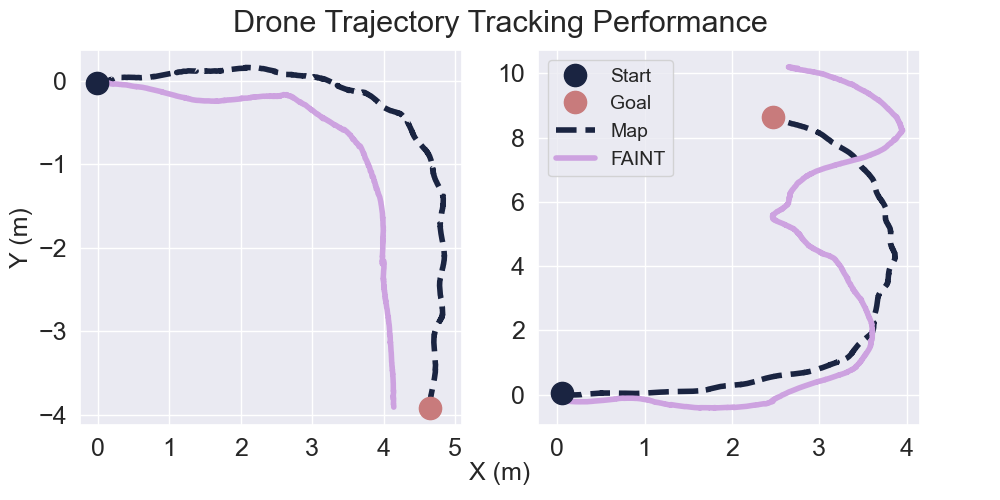 }
    \caption{ Drone trajectory relative to $\map$ when controlled by \MethodName. The drone successfully reached the goal on the left trajectory ($RMSE$ \SI{1.07}{\meter}), but missed it on the right ($RMSE$ \SI{1.42}{\meter}).}
    \label{fig:drone_trajectories}
    \vspace{-0.3cm}
\end{figure*}

To study \textbf{Q2.}, we compared \MethodNameSim with topological navigation methods from previous work, trained with real-world data.
We only considered methods that can run onboard the robot, which rules out larger models such as CrossFormer~\citep{doshi_scaling_2024}. The baseline models were deployed with the author-provided model checkpoints.

We extend the experiment setup of Sec.~\ref{sec:sim_vs_real} with a new route type with illumination change between map collection and deployment. Additionally, we conducted a controlled study on illumination change in which we captured maps on one easy route under regular indoor lighting~\citep{dilaura_lighting_2011} (\SI{220}{\lux}) and dim lighting (\SI{25}{\lux}). During testing, illumination was progressively transitioned between these two levels.

\noindent\textbf{Results} of the SOTA comparison are shown in Table~\ref{tab:sota_experiment}.
NoMAD performs surprisingly poorly. We hypothesize the diffusion-based method to be less robust to variation in cameras and embodiments.
\MethodName outperforms other methods across every route category. The baselines struggle with segments that require sharp turns, often failing to navigate effectively toward the goal. We hypothesize this is caused by the limited receptive fields of their convolutional goal-conditioning modules.
In contrast, \MethodName’s binocular encoder allows it to associate observation and goal image features even under wide baseline changes (see Fig.~\ref{fig:obs_goal_attn}).
The performance differences were the most drastic on routes with illumination change.
The baselines struggle with even moderate changes, while \MethodName's performance is consistent across illuminations.
With controlled illumination change (Fig.~\ref{fig:control_exp_illuminatinons_example}), both ViNT and \MethodName perform well when the deployment illumination is close to the reference, but as the difference grows, ViNT degrades and fails completely.
With the \SI{25}{\lux} map, it only succeeds if the deployment illumination is the same. In contrast, \MethodName succeeds under all conditions with both references.
This robustness is likely due to \MethodName’s use of a pre-trained visual encoder, which helps bridge appearance gaps — including those caused by lighting changes.

\subsection{Simulation-to-real generalization}
\label{sec:sim2real_ablation}
To address \textbf{Q3.} we analyze model design choices' effects on sim2real generalization.
We trained models with $12M$ samples from simulation with either behavior cloning (BC) or DAgger.
Each method was evaluated over 3 repetitions across 10 real-world routes, and on the 2500 routes of the HM3D Val split~\citep{ramakrishnan_habitat-matterport_2021}.

% Architecture ablations
\begin{table}
    \centering
    \caption{\textbf{Simulation-to-real} experiment \emph{success rates} ($\uparrow$). Note that ViNT does not allow freezing the encoder.}
    \label{tab:sim2real_ablations}
    \begin{adjustbox}{width=\linewidth}
    
\begin{tabular}{llcccc}
\textbf{Method} & \textbf{Encoder type} & \makecell{\textbf{Encoder} \\ \textbf{frozen}} & \textbf{Mode} & \makecell{$\mathbf{Real}$ \\ $n=30$} & \makecell{$\mathbf{Sim}$ \\ $2500$} \\ \hline

\textbf{\MethodName} & Theia CDDSV & \checkmark & BC & $0.23$ & $0.87$ \\

\best & Theia CDDSV & \checkmark & DAgger & $\mathbf{0.80}$ & $\mathbf{0.91}$ \\

& Theia CDIV & \checkmark & DAgger & $0.60$ & $\mathbf{0.91}$ \\

& EfficientNet-B0 & \checkmark & DAgger & $0.13$ & $0.79$ \\ \hline

\textbf{ViNT}~\citep{shahViNTLargeScaleMultiTask2023} & EfficientNet-B0 & \ding{56} & DAgger & $0.40$ & $0.87$ \\ \hline
\end{tabular}

    \end{adjustbox}
    \vspace{-0.4cm}
\end{table}
\noindent\textbf{Results} in Table~\ref{tab:sim2real_ablations} highlight the role of image encoder pretraining and on-policy data for sim2real transfer. The frozen EfficientNet~\citep{tanEfficientNetRethinkingModel2019} trained for classification on ImageNet transfers poorly, performing even worse than ViNT with an unfrozen encoder.
The two Theia~\cite{shang_theia_2024} variants distill representations from large models trained for diverse visual tasks, and enable strong real-world performance even without fine-tuning.
Models trained with BC and DAgger perform similarly in simulation but differ greatly in real-world performance. We attribute this to the stronger prediction error compounding caused by the sim2real gap and higher non-determinism of the real world.
DAgger exposes the policy to a wider state distribution during training, drastically improving real-world performance.

\subsection{Cross-embodiment generalization}
To study \textbf{Q4.}, we trained \MethodName with a simulated wheeled robot embodiment and deployed to a real drone without any modifications.
The policy controlled the drone's forward velocity, up to \SI{0.4}{\metre/\second}, and yaw rate at a fixed elevation of $\sim$\SI{1.5}{\metre}. We tested drone navigation on two 'open space'-type routes.
Fig.~\ref{fig:drone_trajectories} shows the drone trajectories and tracking metrics compared to the reference routes.
These preliminary results indicate that the deployment embodiment does not have to be strictly similar to the one used in training. We leave more thorough analysis of cross-embodiment generalization to future work.

%===============================================================================
%\section{Results \& Analysis}
%\label{sec:results}
%\input{tex/5-results}

%===============================================================================
\section{Conclusion}
\label{sec:conclusion}
This work demonstrated that a simulation-trained visual navigation policy can reach performance comparable to policies trained with real data.
We proposed a novel navigation policy architecture that is similar to the previous state-of-the-art, but introduces key modifications that make it suitable for training with both real and synthetic data.
Comparison of synthetic and real data trained versions of the policy show that on-policy learning is a major advantage of simulated training, providing robustness to covariate shift.
The findings suggest combining off-policy real-world datasets with on-policy corrections from simulation as an interesting avenue for future work.

%%%%%%%%%%%%%%%%%%%%%%%%%%%%%%%%%%%%%%%%%%%%%%%%%%%%%%%%%%%%%%%%%%%%%%%%%%%%%%%%


%%%%%%%%%%%%%%%%%%%%%%%%%%%%%%%%%%%%%%%%%%%%%%%%%%%%%%%%%%%%%%%%%%%%%%%%%%%%%%%%

%%%%%%%%%%%%%%%%%%%%%%%%%%%%%%%%%%%%%%%%%%%%%%%%%%%%%%%%%%%%%%%%%%%%%%%%%%%%%%%%

%%%%%%%%%%%%%%%%%%%%%%%%%%%%%%%%%%%%%%%%%%%%%%%%%%%%%%%%%%%%%%%%%%%%%%%%%%%%%%%%
%\section*{APPENDIX}
%\counterwithin{figure}{subsection}
%\counterwithin{table}{subsection}
%\input{tex/supplementary.tex}

\section*{ACKNOWLEDGMENT}

The authors wish to acknowledge CSC – IT Center for Science, Finland, for generous computational resources. The work was financially supported by the Technology Innovation Institute, and also received funding from the European Commission’s HORIZON.1.2 - Marie Skłodowska-Curie Actions (MSCA) under Grant agreement No. 101072634, project RAICAM.

%%%%%%%%%%%%%%%%%%%%%%%%%%%%%%%%%%%%%%%%%%%%%%%%%%%%%%%%%%%%%%%%%%%%%%%%%%%%%%%%

%\def\url#1{}
\bibliographystyle{plainnat_nourl}
\bibliography{references_manual}  % .bib

\end{document}